\documentclass[a4paper]{article}

\usepackage[english]{babel}
\usepackage[utf8]{inputenc}
\usepackage[T1]{fontenc}

\usepackage[a4paper,top=2cm,bottom=2cm,left=3cm,right=3cm,marginparwidth=1.75cm]{geometry}

\usepackage{amsmath}
\usepackage{graphicx}
\usepackage{url}
\usepackage[colorinlistoftodos]{todonotes}
\usepackage[colorlinks=true, allcolors=blue]{hyperref}
\usepackage[rightcaption]{sidecap}
\usepackage{float}
\usepackage[affil-it]{authblk}
\usepackage{amsthm}
\usepackage{graphicx}               % Necessary to use \scalebox
\usepackage{amssymb}
\numberwithin{equation}{subsection}
\usepackage{environ}
\usepackage{bm}

\usepackage[backend=biber,sorting=ynt]{biblatex}

\NewEnviron{myequation}{%
\begin{equation}
\scalebox{1.5}{$\BODY$}
\end{equation}
}

\title{\textbf{A New Approach To Text Rating Classification Using Sentiment Analysis}}
\author{\textbf{Thomas Konstantinovsky}}
    \affil{Department of Computer Science, Holon Institute of Technology}

\begin{document}
\maketitle

\begin{abstract}
Typical use cases of sentiment analysis usually revolve around assessing the probability of a text belonging to a certain sentiment and deriving insight concerning it; little work has been done to explore further use cases derived using those probabilities in the context of rating. In this paper, we redefine the sentiment proportion values as building blocks for a triangle structure, allowing us to derive variables for a new formula for classifying text given in the form of product reviews into a group of higher and a group of lower ratings and prove a dependence exists between the sentiments and the ratings.
\end{abstract}

\providecommand{\keywords}[1]
{
  \small	
  \textbf{\textit{Keywords---}} #1
}

\section{Introduction}
In our current day and age, reviews are part of almost every product/service provided on the internet\cite{inbook}, as seen in \cite{Lackermair2013ImportanceOO} it is the primary way for a company to get an understanding concerning the amount of success their product has and as examined in \cite{trust} for the customer to build trust in purchasing or using a service of which only a description or a picture exits.
Therefore, a need for a deeper understanding and analysis of those reviews are needed\cite{inproceedings} for any individual who wishes to derive various consequences regarding a product.
Standard methods for such insight derivation include sentiment analysis, around which we will formulate a new approach for review rating classification.
Reviews across the internet mainly consist of text-based and rating-based formats, where in many cases, a combination of both is considered a single review; the method developed in this paper focuses on the ability to associate a review to a rating cluster based on sentiment proportions.
We will define two main groups: one group consisting of a majority of reviews higher than three stars (in a 5-star ranking system) and another group of all reviews, which correspond to the less than three stars.

As the analyzing end of a set of reviews, we may not have access to a categorical ranking system provided with the reviews as part of the data.
To solve this problem, we might want to generate ratings or classify the reviews, but this is not a trivial problem as studies suggest that rating inference shall not be tackled as a classical multi-category classification problem, as the ordering of class labels in rating inference is essential \cite{apstr}, reinforced by ambiguity, one of the biggest challenges in natural language processing, elements such as different personalities and sarcasm skew the results of traditional classification models \cite{sarcas_sent}.

Having the review text is enough to calculate and extract the sentiment proportions; using those proportions, we will construct new attributes and a new hyperparameter from those new attributes that will allow us to project the data into a space that separates the reviews into two main groups, as stated before.

Figure 1 shows the spread of different tripadvisor hotel reviews\cite{barkha_bansal_2018_1219899} plotted via their sentiment proportions; the color scale represents the rating given in each of the reviews; throughout our calculations, the real rating that the person left is only for visual conformation and is never included in any part of our calculation.
There is no simple solution when considering separating the reviews with high ratings (points that are closer to yellow color) from the reviews with low ratings (points that are closer to a blue color) without using any machine learning process or including the actual rating values in a calculation.
  
\begin{figure}[H]
\caption{Example of a Review Data-set Plotted Using Sentiment Values }
\centering
\includegraphics[width=0.9\textwidth]{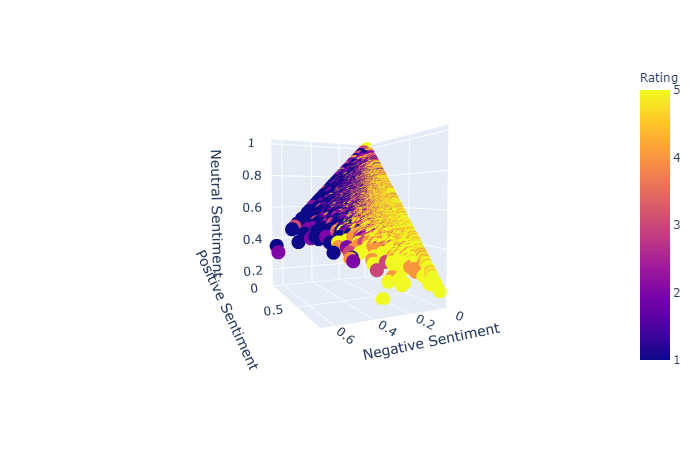}
\end{figure}

A natural assumption to be made is of dependence between a review text and its corresponding rating given by the user; following this assumption, one can associate a review to a rating based on the text alone and as the case in this paper using the text's sentiment proportions.
In contrast to the topic research in \cite{HU201442} where a connection between the sentiment, rating and their inner relation to costumer decision making was examined, the purpose of this paper is to establish a direct dependence and define the rating as a function of a review sentiments.
The equation proposed gives an approximation for such a separation without iteratively training any model, a separation that helps us generate ratings and confirms our assumption.

\section{Methods}

\subsection{The Vader Algorithm}
The building blocks from which we derive our formula are generated using The Vader sentiment analysis algorithm \cite{conf/icwsm/HuttoG14}, which takes text as an input and returns three values, which we will refer to throughout the paper as "Pos" - the proportion of sentences in the text that are tagged positive "Neg" - the proportion of sentences in the text that are tagged negative and "Neu" = the proportion of sentences in the text that are tagged neutral.

\theoremstyle{definition}
\newtheorem{definition}{Definition}
\begin{definition}
$$0\leq(Pos,Neu,Neg)\leq1$$
$$Pos+Neu+Neg = 1$$
\end{definition}

\subsection{Translation Into A Triangle}
The first step towards our formula consists of deriving attributes from a triangle.
The triangle from which we will derive those attributes will be constructed using the sentiments extracted using the Vader algorithm; for us to construct such a triangle, we transformed the 1D values into 2D vectors to place them on 3 different line spans.
A 2-dimensional space was divided into three where the angles between each dividing line are 120 degrees; each dividing line represents the vector on which the values (Pos/Neg/Neu) extracted by the Vader the algorithm will span, as can be seen in Figure 2.

\begin{figure}[H]
\caption{The Line On Which Each Sentiment Will Span}
\centering
\includegraphics[width=0.5\textwidth]{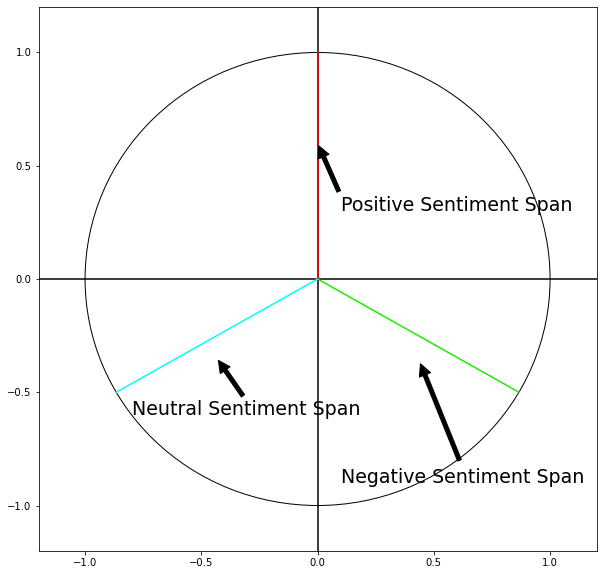}
\end{figure}

The three sentiment values are multiplied by rotation vectors to receive new points located on the appropriate span line as we defined above.
This transformation $\mathbb{R}^1\longrightarrow \mathbb{R}^2$ is defined as follows;
\theoremstyle{definition}
\begin{definition}
$$\bm{Pos} = [0,Pos]$$
$$\bm{Neg} = Neg\cdot[cos(\frac{-\pi}{4}),sin(\frac{-\pi}{4})]  $$
$$\bm{Neu} = Neu\cdot[cos(\frac{-5\pi}{4}),sin(\frac{5\pi}{4})]  $$
\end{definition}

\subsection{Formulating Variables}

Now that Pos, Neg, and Neu are vectors in $\mathbb{R}^{2}$ spanning on the lines already defined, we derive the variables needed for our equation.

\theoremstyle{definition}
\begin{definition}[a, b and c variables representing triangle side lengths ]
$$a =||\bm{Neu} - \bm{Pos}||$$
$$b =||\bm{Neu} - \bm{Neg}||$$
$$c =||\bm{Pos} - \bm{Neg}||$$

\end{definition}

\theoremstyle{definition}
\begin{definition}[$\alpha$ is calculated using Heron's formula\cite{heron}]
$$s = \frac{(a+b+c)}{2}$$
$$\alpha = \sqrt{s\cdot (s-a) \cdot (s-b) \cdot(s-c)}$$

\end{definition}

\begin{definition}[$\beta$ is the inner product weighted by the magnitude ratio between the difference from $\bm{Neu}$]
$$\bm{u}= \bm{Neu} - \bm{Pos} \;\;\; \bm{v}= \bm{Neu} - \bm{Neg}$$
$$\beta = \arccos((\bm{u}\cdot \bm{v})\cdot \frac{||\bm{v}||}{||\bm{u}||})$$
\end{definition}

\theoremstyle{definition}
\begin{definition}[$\gamma$ is the height of a triangle where the maximum side length is considered to be the base and $\alpha$ to be the area of the triangle]
$$T_{h} = max(a,b,c)$$
$$\gamma =  \frac{2\cdot \alpha }{T_{h}} $$
\end{definition}

\subsection{The $\Omega$ hyper-parameter}

After deriving the variables, we can now formulate the equation for our new hyperparameter.
In the process of constructing the formula, different functions of Pos, Neg, and Neu were tested while keeping in mind the goal of creating a sufficient separation between rating clusters.
No significant results were achieved; this called for formulating a new hyperparameter dependent on variables defined in the previous section, representing the separation dimension.
The equation's general form resulted from a conceptual idea translated into mathematical notations, and we refer to this hyperparameter as $\Omega$.

\theoremstyle{definition}
\begin{definition}

\begin{myequation}%
\Omega = \ln(\alpha^{\beta}+\epsilon) \% e^{\gamma} \;\;|\;\; \epsilon  \longrightarrow 0 %
\end{myequation}

\end{definition}

The idea behind the structure of the formula is finding a meaningful rate of growth for a sequence of rational numbers that converge to the irrational exponent; conceptually, the purpose of the natural logarithm is to retrieve the number of units or the “time” at which a sequence of $\alpha$'s converges to the $\beta$'s as defined in the previous section.
Afterward, we take the remainder (modulus) of the “growth rate” with the $\gamma$ variable’s growth value; In other words, the hyperparameter should represent different states of the growth rate in terms of the exponent.

Each review will be now represented as a vector in $\mathbb{R}^3$ where the vector elements are : $[a,c,\Omega]$.
Projecting each review vector to such a space creates a clear separation between two groups of different rating consistencies one consists of a majority of high ratings i.e., higher than 3 out of 5, and the second of those with lower ratings i.e., lower than 3 out of 5.\newline

\section{Results}
We have observed that by looking at the spread of data in 3 dimensions where our axes were $\Omega$ which is the new hyperparameter calculated, “a” and “c” we get a segmentation into 2 main clusters, a cluster with a majority of highly rated samples and a cluster of samples with a majority lower ratings.

\begin{figure}[H]{\cite{plotly}}
\caption{Resulting Data Spread - Hotel Reviews}
\centering
\includegraphics[width=0.9\textwidth]{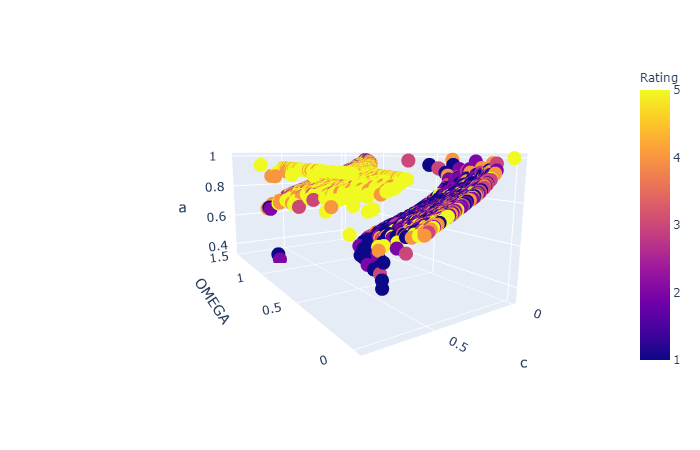}
\end{figure}

The same results were archived on multiple different datasets with different ranges of themes and topic; as stated in the introduction, the main example shown throughout the paper is of Tripadvisor hotel reviews, figure 4 and figure 5 shows the same process applied to a dataset of different earphones reviews from Amazon\footnote{https://www.kaggle.com/shitalkat/amazonearphonesreviews}, and a dataset of wine reviews\footnote{https://www.kaggle.com/krrai77/wine-reviews}.

\begin{figure}[H]
  \centering
  \begin{minipage}[H]{0.4\textwidth} 
    \includegraphics[width=\textwidth]{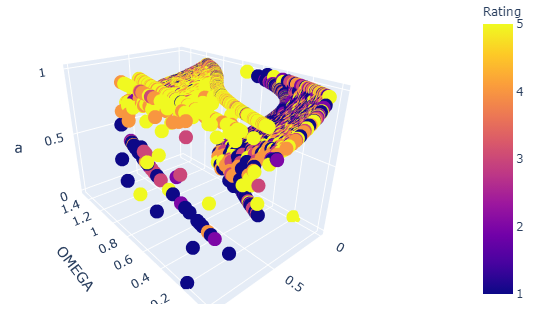}
    \caption{Resulting Data Spread - Amazon Headphone Reviews}
  \end{minipage}
  \hfill
  \begin{minipage}[H]{0.4\textwidth} 
    \includegraphics[width=\textwidth]{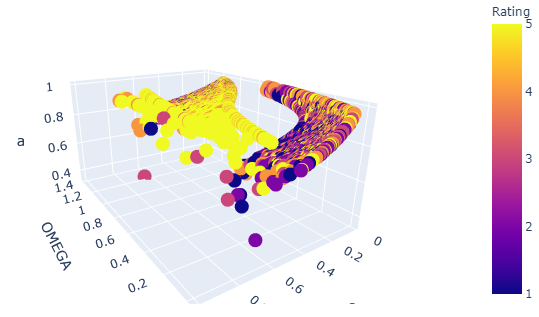}
    \caption{Resulting Data Spread - Wine Reviews}
  \end{minipage}
\end{figure}

\subsection{Cluster Extraction}
Extracting the clusters formed is easily done using the DBSCAN algorithm \cite{ester1996densitybased} where the optimal clustering as tested on multiple datasets, including outlier detection, is achieved by using a minimum neighborhood size of 7 and an epsilon radius of 0.09.

\begin{figure}[H]
\caption{Clusters Tagged Using DBSCAN}
\centering
\includegraphics[width=0.9\textwidth]{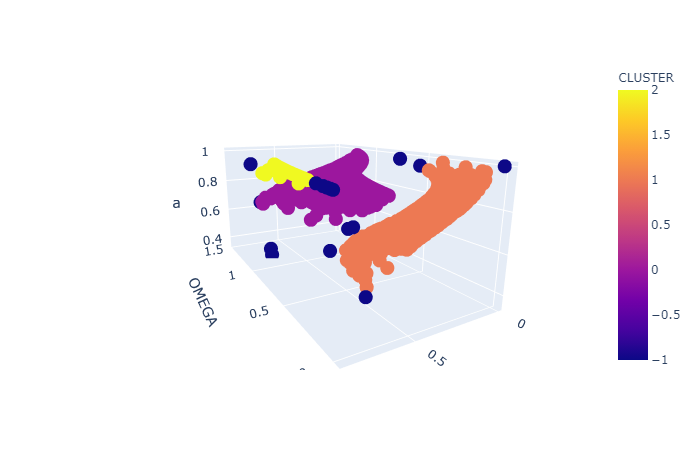}
\end{figure}

Figure 6 shows our formula's resulting separation, the negative cluster, which represents the low rated reviews, i.e., under 3 out 5 stars is colored in orange.
The positive cluster representing the high-rated reviews, i.e., higher than 3 out 5 stars, is colored in purple.
The most positive reviews, strictly 4 and above stars rated, in the dataset are colored yellow, represented by the line above the positive cluster; note that in a case where there are extremely negative reviews, they will appear similarly as a line under the negative cluster.
Outliers were colored in blue as they deviate extremely from the main behavior of the remaining points.
 
\newpage

\begin{figure}[H]
\caption{Our Clusters in Terms of The Initial Axes, Similarly to Figure 1}
\centering
\includegraphics[width=0.6\textwidth]{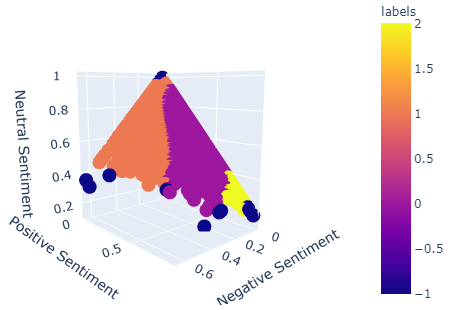}
\end{figure}

In Figure 7, the clusters tagged using DBSCAN in the transformed space were colored using the same colors to emphasize the non-linear separation achieved without iteratively training any model or searching for some separating margin.
Not only do we see how our main clusters capture the regions of rating above and under 3-stars as mentioned before, but we also can see how the yellow line that we saw in Figure 6 and the blue outliers are located with respect to the entire dataset, the blue outliers are on the extremes of each vertex of the triangular spread of points and the yellow cluster, i.e., "most positive." ratings which are strictly  4 and above stars is on the extreme vertex of low negative sentiment and high positive sentiment.

\subsection{Cluster Characteristics}
To get a better understanding of the characteristics of the clusters and how ratings are distributed within them, Bayesian inference was used to understand what are the underlying probabilities of the cluster ratings,
To generate such a process, a Monte Carlo Markov Chain algorithm \cite{Walsh04markovchain} was used to sample a posterior space for a vector $\bm{\Theta}$.$$\bm{\Theta} = [\theta_{1},\theta_{2},\theta_{3},\theta_{4},\theta_{5}]$$ It is modeled using a Dirichlet distribution\cite{2011} initiated with a uniform prior as we do not want to assume any prior knowledge regarding the rating probabilities in each cluster.
$$\alpha = [1,1,1,1,1]$$
$$\bm{\Theta} \sim Dir(\alpha)$$
The Theta probability vector will be the input for a multinominal model from which we will sample our probabilities for each rating.
$$N = \text{number of samples}$$
$$P_{Rating} \sim Multinominal(N,\bm{\Theta})$$

It is important to note that we hypothesize that in the case of a bad separation, both of the major clusters we saw in the previous section will have close to a uniform distribution for each rating probability, i.e., each rating will have a probability of around 20\% of appearing in the cluster.
In contrast, if indeed the separation was significant, we expect to see the positive cluster, which is colored in purple in figure 4, to have most of its credibility associated with ratings higher than 3, the same goes for the negative cluster colored orange in figure 4 which we expect to have most of its credibility associated with ratings lower than 3.

\begin{figure}[H]
\caption{Positive Cluster Posterior Rating Probability Distribution}
\centering
\includegraphics[width=0.9\textwidth]{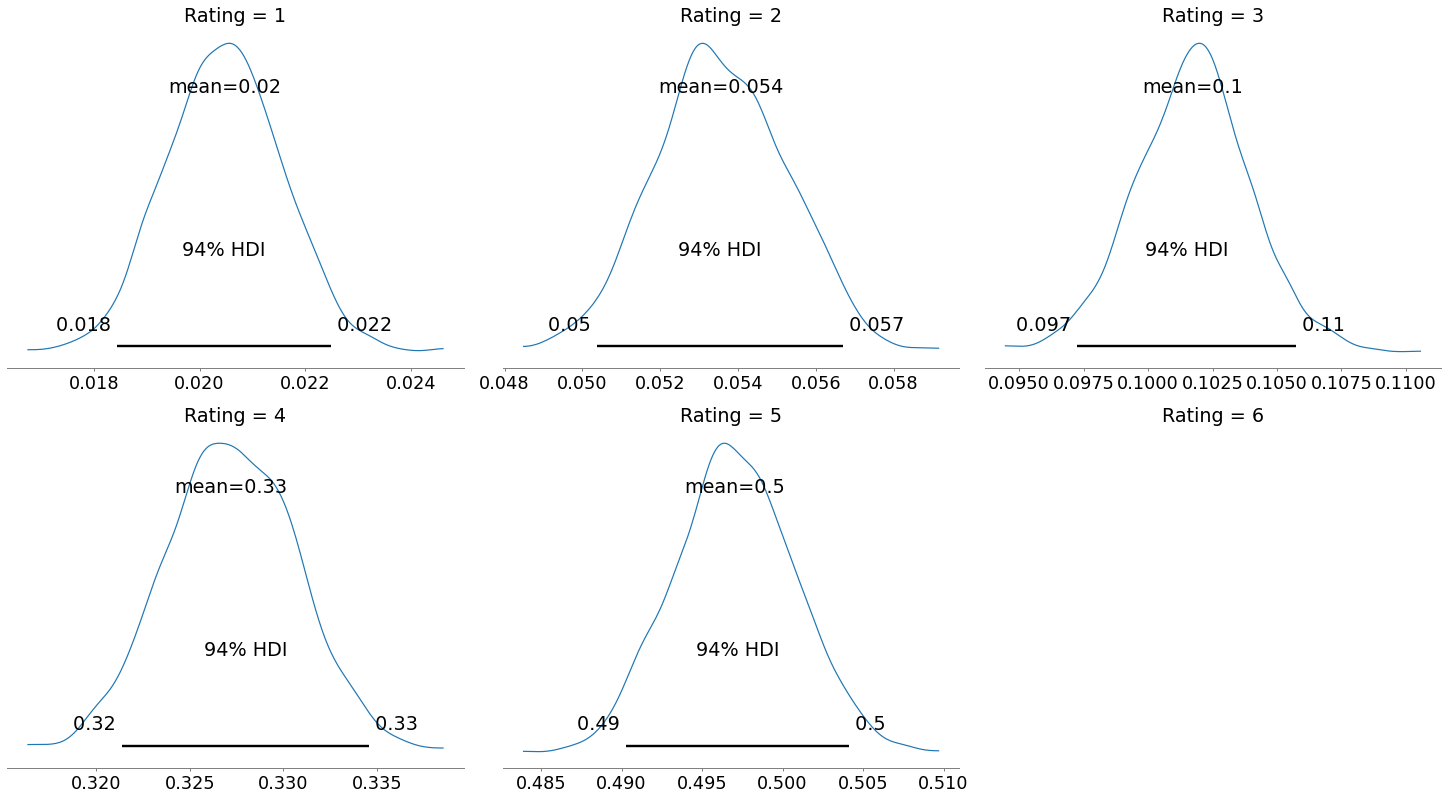}
\end{figure}

As can be seen in Figure 8, when we inferred the positive cluster as described above, it is clear a 5-star rating is the most probable when considering a random text that belongs to the positive cluster; in fact, a text belonging to this cluster has 83\% chance of being higher than 4 stars when taking in mind the anomalies mentioned earlier in the paper such as users who leave a positive review with a low rating will appear in this cluster.
Another important notice is that the posterior distributions' deviation is quite tight around a small interval, a fact that supports our hypothesis and reinforces our confidence in the probability estimates.

\begin{figure}[H]
\caption{Negative Cluster Posterior Rating Probability Distribution}
\centering
\includegraphics[width=0.9\textwidth]{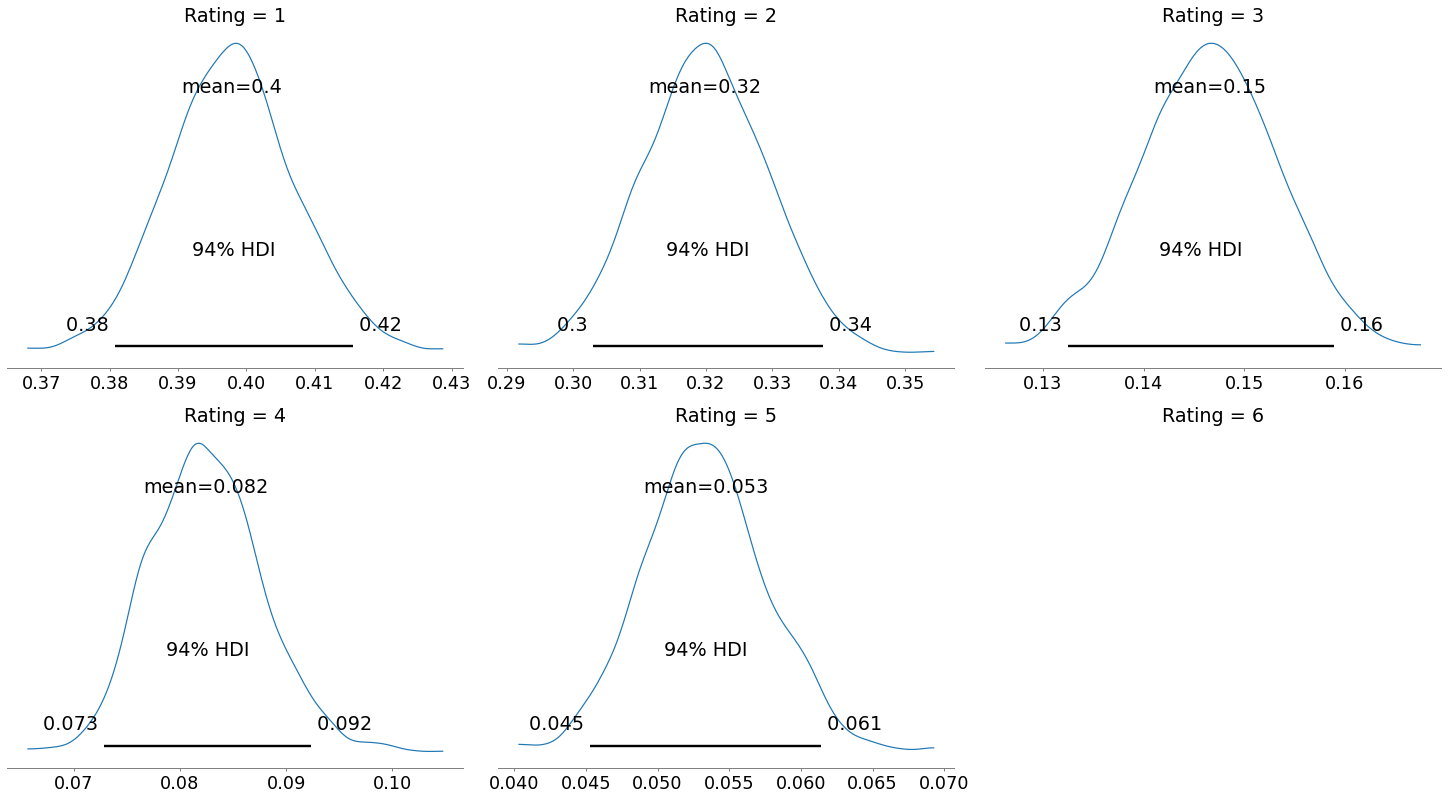}
\end{figure}
Similarly to the interpretation, we did in Figure 8, Figure 9 represent the same process applied to the negative cluster, and we see that the results in this cluster support our hypothesis as well, not only the highest probability for a rating of a text belonging to this cluster is 1-star, there is 87\% chance that a point from this cluster will have a rating lower than 3-stars. Similarly to the previous explanation, there may be positive reviews with low ratings, which will appear in the negative cluster.

\section{Discussion}
At first, when looking at the distribution of the reviews sentiments and the rating behind each point, as seen in Figure 1, it appeared that there might be a relationship between the sentiment proportion given by the Vader algorithm and the rating, we hypothesized that the reviews with lower rating could be naturally clustered so as the review with higher ratings, an equation for a new hyperparameter was introduced, thanks to this new hyperparameter we achieved a transformation in which the data is separated into clusters that meet our hypothesis as we explained in the previous section.
Empirically proving the dependence between the rating and the sentiment proportions and formulating the equation in this paper not only provides a new tool to the field of natural language processing for generating ratings based only on text sentiments in the absence of such data but also proposes a further investigation into the given equation and the application it might have on different fields.
The process can be repeated for any vector of values meeting the criteria given in Definition 1.

The anomalies seen within the clusters raise a question about the effect of the prior distribution of ratings on new reviews being written at a certain platform; we might think of the rating given by a user in an analogous way to the Chinese restaurant stochastic process\cite{10.1007/BFb0099421}, where we assume that each new review rating is affected directly by the current total ratings, i.e., a person is more likely to leave a high rating to a product/service that is highly rated even if he thinks lower of the product and vice versa when the total rating of the product/service is low.
When we inferred the clusters, we saw that there were lower ratings residing in the positive cluster and higher ratings residing in the negative cluster; in the same manner, as in the Chinese restaurant process, a new guest has a probability of choosing a table that is not strictly the one with the highest probability we see people who are giving a rating opposite to the sentiment of their review.
In future studies, it will be of value to give an analytic proof for the equation present in this paper.

\section{Conclusions}
A new formula was introduced in this paper, enabling deeper analysis of text data in the form of reviews by proposing a direct calculation for rating classification into one of two clusters.
Not only did we establish a dependence between the sentiment proportions of a review to the corresponding rating, but we also achieved a direct transformation from the sentiment proportions to the separated space without any use of classic iterative converging machine learning algorithms.

\medskip

\printbibliography

\end{document}